\documentclass{article}
\usepackage{multicol}
\usepackage{multirow}
\usepackage{bcprules}
\usepackage[T1]{fontenc}
\usepackage{graphicx}
\usepackage{hyperref}

\usepackage{url}            
\usepackage{amsfonts}       
\usepackage{float}

\usepackage{amsmath,amssymb}
\usepackage{mathtools}      
\usepackage{multicol}

\newcommand{\TRUE}{\mathtt{true}}
\newcommand{\FALSE}{\mathtt{false}}
\newcommand{\AND}{\mathtt{and}}
\newcommand{\OR}{\mathtt{or}}
\newcommand{\NOT}{\mathtt{not}}
\newcommand{\Lb}{\mathrm{L}}

\newcommand{\rrightarrow}{\mathrel{\mathrlap{\rightarrow}\mkern1mu\rightarrow}}
\newcommand{\betared}[0]{\rightarrow_\beta}
\newcommand{\fullbetared}[0]{\rrightarrow_\beta}

\usepackage{amsthm}
\newtheorem{definition}{Definition}
\usepackage{authblk}

\title{Towards a Neural Lambda Calculus: Neurosymbolic AI Applied to the Foundations of Functional Programming}

\author{Jo\~{a}o M. Flach, \'{A}lvaro F. Moreira, Lu\'{i}s C. Lamb}

\affil{Institute of Informatics \\
 Federal University of Rio Grande do Sul\\
 Porto Alegre, RS, Brazil\\
  jmflach@inf.ufrgs.br, alvaro.moreira@inf.ufrgs.br, luislamb@acm.org}

\begin{document}
\maketitle

\begin{abstract}
In recent decades, deep neural network-based models have become the dominant paradigm in machine learning. The use of artificial neural networks in symbolic learning has recently been considered increasingly relevant. To study the capabilities of neural networks in the symbolic AI domain, researchers have explored the ability of deep neural networks to learn mathematical operations, logic inference, and even the execution of computer programs.
The latter is known to be too complex a task for neural networks. Therefore, the results were not always successful and often required the introduction of biased elements in the learning process, in addition to restricting the scope of possible programs to be executed.
In this work, we will analyze the ability of neural networks to learn how to execute programs as a whole.
To do so, we propose a different approach. Instead of using an imperative programming language with complex structures, we use the Lambda Calculus ($\lambda$-Calculus), a simple and Turing-complete mathematical formalism, which serves as the basis for modern functional programming languages and is at the heart of computability theory. We will introduce the use of integrated neural learning and lambda-calculi formalization. We explore that the execution of a program in $\lambda$-Calculus is based on reductions, and we will show that it is enough to learn how to perform these reductions so that we can execute any program. \footnote{This manuscript is an improvement of a previous one posted in arXiv.org by Flach and Lamb, with significant contributions by Professor A.F. Moreira.}\\

\end{abstract}


\section{Introduction}
\label{sec:intro}

In machine learning, there has been a long-standing debate about the best way to approach the task of learning from data \cite{lecunhintonbengio}. Rule-based inference emphasizes the use of explicit logical rules to reason about the data and make decisions, inferences, and predictions \cite{garcez2020neurosymbolic}. The other perspective,  statistical learning, involves using mathematical models to automatically extract patterns and relationships from data \cite{rumelhart86a,lecunhintonbengio}.

Deep neural networks have been used successfully in
applications such as speech recognition, machine translation, and handwriting recognition. Recently, advances in the field have resulted in the introduction of models that are changing the landscape and allowing us to tackle a wider range of problems, including symbolic ones, using neural networks. When neural networks are applied to symbolic problems, the result is a hybrid approach that combines the advantages of both rule-based (in which logic-based ones are prominent) and statistical approaches. This combination falls into the realm of \textit{neurosymbolic AI} \cite{Garcez2009,kautz2022third}. This field combines the statistical nature of machine learning with the logical nature of reasoning in AI \cite{Garcez2009}, and has recently attracted the attention of researchers from several subfields of AI and computer science \cite{garcez2020neurosymbolic,besold2022neural} as it can contribute to offering explainable machine learning approaches. This interest is sparked by the need to build more robust AI models \cite{dietterich2017steps}, as initially advocated by Valiant and now a subject of increasing interest in AI research \cite{valiant2003,valiant2018,marcus2019rebooting}.

In this work, we intend to explore the capacity of machine learning models, specifically the Transformer \cite{Vaswani2017}, to learn rules to perform computations, a task traditionally seen as too complex for neural networks to handle. 


The idea of training a machine learning model to perform computations is relatively new. Most works in this field restrict the class of programs that the model can take as input and consist of teaching the model to understand rules for the evaluation of mathematical expressions.

We, on the other hand, use the Lambda Calculus ($\lambda$-Calculus) as the underlying framework \cite{barendregt1984lambda}. Introduced by Alonzo Church in the 1930s~\cite{church1936unsolvable}, it is a simple, elegant, and Turing-complete formal system based on function abstraction that captures the notion of function definition and function application. It is the base for modern functional programming languages, such as OCaml, Racket, Haskell, and others \cite{michaelson2011introduction}, and it has become one of the main computational models alongside Turing Machines \cite{Hindley2006}.

In essence, the $\lambda$-Calculus can be seen as a programming language whose programs, called $\lambda$ terms, can be subject to reductions that can be interpreted as computations performed within the formalism. Applying a single reduction to a term represents a one-step computation in the $\lambda$-Calculus. On the other hand, a full computation involves successive single reductions of a term until a \textit{ normal form} is obtained (if it has one), i.e., a term that cannot be reduced any further. 

In summary, our research question is: \textit{``Can a machine learning model learn to perform computations?''}. To gradually enhance our understanding of the subject matter and enable us to provide an answer to our research question, we propose the following two hypotheses. 
\begin{itemize}
    \item \textit{H1: The Transformer model can learn to perform  one-step computation on Lambda Calculus.}
    \item \textit{H2: The Transformer model can learn to perform  full computation on Lambda Calculus.}
\end{itemize}

Considering that lambda terms do not have a fixed size, we use a sequence-to-sequence (seq2seq) model, which can take inputs and produce outputs of any length. Specifically, we use a model that has been widely used for several types of applications and has also been tested for symbolic tasks: the Transformer \cite{Vaswani2017}.

\section{Related Work}

In \cite{zaremba2014learning}, seq2seq models are used to learn to evaluate short computer programs using an imperative language with the \textit{Python} syntax. However, their domain is restricted to short programs that can use just some arithmetic operations, variable assignment, if-statements, and for loops (not nested). 

Some other studies also have worked towards developing models that learn algorithms or learn to execute computer code, including \cite{kaiser2015neural,graves2014neural,trask2018neural}. However, the domain of these works is restricted to some arithmetical operations or sequence computations (copying, duplicating, sorting).
Additional work concentrates on acquiring an understanding of program representation. For example, \cite{maddison2014structured} builds generative models of "natural source code", i.e., code written by humans and meant to be understood by other humans, while \cite{mou2014building} applies neural networks to determine if two programs are equivalent.

The Transformer model \cite{Vaswani2017} brought several key advancements and improvements compared to the state-of-the-art seq2seq models prevalent at that time. This new model boasted improved parallelism, reduced sequential processing requirements, and the ability to handle longer sequences, among other things. These innovative features have contributed to the widespread adoption of the Transformer model in various Machine Learning applications.

 In a study by \cite{lample2019}, the Transformer model was applied to learn how to symbolically integrate functions, producing very promising results. The authors demonstrated that the model was capable of learning to perform integrals in a way that was both accurate and efficient, outperforming existing methods in many cases. This study highlights the versatility and potential of the Transformer model, making it a valuable tool for tackling a wide range of machine learning tasks, especially in areas that require symbolic reasoning and mathematical operations.

In addition, the Transformer model has enabled recent developments in chatbot technologies. An example of a chatbot that has emerged as a result of this development is the \textit{ChatGPT} \footnote{available at: \url{https://openai.com/blog/chatgpt/}}, which is based on a state-of-the-art AI model, the GPT-3, from \cite{Brown2020}. These chatbots can answer questions about various subjects \cite{Roose2022} and perform basic symbolic reasoning. However, their symbolic reasoning capability is still limited, giving some incorrect answers to straightforward questions.

In the present work, we shift from the imperative paradigm that all other works have used, to the functional paradigm, which is the case for the $\lambda$-Calculus. With this, we abstract the idea of learning to compute computer programs to learn to perform reductions in $\lambda$ terms. 

With this approach, in the sequel we will show that to learn one-step computation, we obtained an accuracy of $88.89\%$ in completely random terms $\lambda$ and of $99.73\%$ in terms representing Boolean expressions. For the full computation task, we obtained an accuracy of $97.70\%$ for $\lambda$ terms that represent boolean expressions. Taking into account a string similarity metric, most of our predicted $\lambda$ terms were above $99\%$ similar to the correct ones.

With these results, we think that the change to the functional paradigm and the use of the Transformer model are two improvements that will be relevant in future research. 

\section{The Lambda Calculus: A Summary}\label{sec2}

In this section, we present an overview of the main aspects of the Lambda Calculus that will be used in the paper.  

\subsection{Syntax} \label{sec:lambda_syntax}

We start by defining the syntax of lambda terms. In the following, $x$ denotes a member of an infinite and countable set of variables. 

\begin{definition}[$\lambda$-terms]
\label{def:grammar}
The set of $\lambda$-terms is defined by the following abstract grammar:
\[
\begin{array}{lcl}
M, N, \ldots  ::= &  x ~~|~~ \lambda x. M ~~|~ ~ M ~N
\end{array}
\]
\end{definition}

Every variable is a $\lambda$ term; the term $\lambda x . M$ is a function with parameter $x$ and body $M$; the term $M\; N$ is the application of the term $M$ to the argument $N$. 

In a term $\lambda x. M$, all the occurrences of variable $x$ inside $M$ are called \textit{bound occurrences}. 

\subsection{Reductions}

 The main reduction of the formalism is a binary relation between terms called $\beta$ reduction, which is based on the substitution operation. A substitution takes a $\lambda$ term and substitutes a variable in it with another $\lambda$ term, similar to what is done in mathematics when a function is applied to an argument. 

%
\begin{definition}[\textit{Redex}] A \textit{redex} (reducible expression) is any subterm in the format 
\[
    (\lambda x.M)\; N
\]
\end{definition}

If a term has no redexes, the term is a normal form. Otherwise, the term is reducible. Now, the $\beta$-reduction relation can be defined as:

\begin{definition}[One-Step]
\label{def:one-step-rules}
$\beta$-reduction ($\betared$) is the smallest relation on lambda terms such that

\small
\begin{multicols}{3}
\infrule{M \betared M'}{M \; N \betared M' \; N}
\infrule{N \betared N'}{M \; N \betared M \; N'}
\infrule{M \betared M'}{\lambda x .M \betared \lambda x .M'}
\end{multicols}
\infax{(\lambda x .M) \; N \betared M[x := N]} 
\normalsize
\end{definition}

In the above definition, $M[x:=N]$  is the notation used to represent the term that results from the substitution of all the occurrences of the variable $x$ in the term $M$ by a term $N$.

The multi-step reduction $\fullbetared$ is defined as the reflexive and transitive closure of $\betared$, as follows:

\begin{definition}[Multi-Step]
\label{multi-step-rules}
$\fullbetared$ is the smallest relation on lambda terms such that
\small

\begin{multicols}{3}

\infrule{}{M \fullbetared M}

\infrule{M \fullbetared N ~ N \fullbetared P}{M \fullbetared P}

\infrule{M \betared N}{M  \fullbetared N}
    
\end{multicols}
\normalsize

\end{definition}

 We say that a term $M$ has a normal form if there is a term $N$ such that $M \fullbetared N$ and N is a normal form.  Not every term has a normal form. One example is the term $(\lambda x. x\; x) \; (\lambda x. x\; x)$ that $\beta$-reduces to itself.

A term can have more than one redex, which means that when we try to apply $\beta$-reduction on a term, we have multiple possibilities. It is useful to have a strategy to select which redex we want to reduce at each computation step. An evaluation strategy is a function that chooses a single redex for every reducible term.

The two most common evaluation strategies are: (i) \textit{normal or leftmost order}, where  the leftmost, outermost redex of a term is reduced first, and the (ii) \textit{applicative or strict order}, where rightmost, innermost redex of a term is reduced first. An outermost redex is a redex not contained in another redex, and an
innermost redex is a redex that does not contain other redexes.

Choosing the evaluation strategy is important to clearly define which redex to reduce through $\beta$-reduction. Furthermore, it is not just a matter of personal preference since there is a theorem that says that if a term $M$ has a normal form $P$, then the \textit{normal order evaluation strategy} will always reach $P$ from $M$, in a finite number of $\beta$ reductions. 

Therefore, in this work, we always use the \textit{normal order evaluation strategy} when making $\beta$ reductions on terms to ensure that if the term has a normal form, we can reach it.

There is another type of reduction in the Lambda calculus, the $\alpha$-reduction, which is responsible for the renaming of bound variables when necessary. But since we are following the Barendregt convention \cite{barendregt1984lambda}, which states that the name of the bound variables will always be unique, we do not need to consider $\alpha$-reduction in this work. 

\subsection{Encodings and Computations}

 The notion of encoding is well-known in Computer Science; for example, our modern computers operate on binary code, on top of which we build abstract ideas. In Lambda calculus, the idea is the same. We can use the structure of function abstractions and applications to encode representations for numbers, booleans, strings, etc.  
 
 In this work, we adopt the Church Encoding for representing Boolean expressions. Examples of such encoding are given below:

%
\small 
\[
\begin{array}{lcl}
\TRUE & \equiv & \lambda x. \lambda y.x\\
\FALSE & \equiv & \lambda x.\lambda y.y\\
\AND & \equiv & \lambda p.\lambda q.~(p ~q) ~p\\
\OR  & \equiv & \lambda p.\lambda q.~(p ~p) ~q\\
\NOT & \equiv & \lambda p.(p ~\FALSE) ~ \TRUE
\end{array}
\]
\normalsize 
With the Boolean constants $\TRUE$ and $\FALSE$, and with the Boolean operator $\AND$, we can codify other Boolean expression such as $\mathit{true ~ \wedge   ~ false}$ as the lambda term $(\AND\ \TRUE) ~ \FALSE$ that reduces to its normal form $\FALSE$ in four small-steps  as follows:
\small
\[
\begin{array}{lcl}
&        &  (\AND\ \TRUE) ~ \FALSE\\
& \equiv &  ((\lambda  p. \lambda   q.~ (p ~ q) ~p) ~ \TRUE) ~ \FALSE\\
& \betared & (\lambda   q.~ (\TRUE\ ~ q) ~\TRUE) ~ \FALSE\\
& \betared & (\TRUE\ ~ \FALSE) ~\TRUE\\
& \equiv   &  ((\lambda  x. \lambda  y.~x) ~ \FALSE) ~ \TRUE \\
& \betared &  (\lambda  y.~\FALSE) ~ \TRUE \\
& \betared &  \FALSE\\
& \equiv   & \lambda x.\lambda y. y
\end{array}
\]

\normalsize

\subsection{Prefix Notation}

The lambda terms presented in the previous section are written in what we call traditional notation. For this work, we only consider $\lambda$-terms in the prefix notation for the learning tasks.  To write lambda terms in the prefix notation, we need to add the application symbol (\textit{@}) to the syntax. An application written as $(N ~M) ~P$ in the traditional notation, for example, is written as $\textit{@} \textit{@} M ~N ~P $ in the prefix notation.

We also change the $\lambda$ symbol for this representation using the uppercase letter ``L''. The terms $\TRUE$, $\FALSE$, and $\AND$ in the prefix notation are:
\small
\[
\begin{array}{lcl}
\TRUE & \equiv &  \Lb ~ x ~  \Lb ~ y  ~x\\
\FALSE & \equiv & \Lb ~ x ~  \Lb ~ y  ~y\\
\AND & \equiv & \Lb ~ p ~ \Lb ~ q ~@ ~@ ~ p~ q ~ p\\
\end{array}
\]
\normalsize 
We chose prefix notation because it offers a well-organized structure derived from a tree-like representation. This structure allows for a more straightforward representation of expressions as it is unambiguous and eliminates the need for parentheses. This makes it easier to process expressions, particularly for the purposes of learning and understanding complex mathematical concepts.

\subsection{De Bruijn Index} \label{sec:de_bruijn}

The De Bruijn index is a tool for defining $\lambda$-terms without having to name the variables \cite{de1972lambda}. 
This approach can benefit us, since the terms are agnostic to the variable naming and are simpler in the sense that they are shorter.  

Basically, it just replaces the variable names by natural numbers. The abstraction no longer has a variable name, and every variable occurrence is represented by a number, indicating at which abstraction it is binded. These nameless terms are called \textit{de Bruijn terms}, and the numeric variables are called \textit{de Bruijn indices} \cite{pierce2002types}. 

For simplicity, we denote the free variables with the number $0$, and the indices of the bound variables start at $1$. This notation assumes that each \textit{de Bruijn index} corresponds to the number of binders (abstractions) under which the variable is. 

This notation can also be used in conjunction with the prefix notation. We will use it to compare with the traditional notation and see if there is any advantage in using a notation with no variable names for the tasks we are interested in.

Here we have examples of lambda terms in de Bruijn notation next to their prefixed notation:
\small
\[
\begin{array}{lclll}
\TRUE & \equiv & \lambda . \lambda . 2  & - &  \quad  \Lb ~ \Lb ~2\\
\FALSE & \equiv & \lambda.\lambda. 1  & - & \quad  \Lb ~ \Lb ~ 1\\
\AND & \equiv & \lambda.\lambda.~(2 ~1) ~2 & - & \quad   \Lb ~ \Lb ~\textit{@} ~ \textit{@} ~2 ~1 ~2 \\
\end{array}
\]
\normalsize

\section{ML and Neurosymbolic AI} \label{sec3}

Machine learning (ML) is a subfield of artificial intelligence that involves the development of algorithms and models that can learn from data to make predictions or decisions. ML algorithms can be trained on large amounts of data, allowing them to identify patterns and relationships in the data and improve their accuracy over time \cite{Goodfellow2016}.

In supervised learning, algorithms are trained on labeled data, where the output or target variable is known \cite{Goodfellow2016}. These algorithms can make predictions about new and unseen data and can be used in applications such as image classification or stock price prediction, for instance.

Unsupervised learning algorithms are trained on unlabeled data where the output or target variable is not known. These algorithms can identify patterns and relationships in the data and can be used in applications that require grouping data into clusters or detecting anomalies in them.

Reinforcement learning algorithms are designed to learn from interactions with an environment. These algorithms receive a reward or penalty for each action they take, and they can be used in various applications, such as game-playing and robotics. 

This work focuses on supervised learning, particularly on connectionist AI (neural networks).

\subsection{Neurosymbolic AI}

Neurosymbolic AI is a field of artificial intelligence that combines the strengths of both symbolic AI and connectionist AI. Symbolic AI represents knowledge in a structured and human-readable form and uses rule-based reasoning systems to perform tasks. Connectionist AI, on the other hand, represents knowledge as patterns in a network of simple processing units. Neurosymbolic models aim to merge the two approaches by incorporating symbolic reasoning and/or representation with neural networks' learning and generalization capabilities \cite{Garcez2009}.

 The paper \cite{kautz2022third} presents six different forms of neurosymbolic AI, varying in how and where the two approaches are combined. In the present work, we use the form \textit{Neuro: Symbolic $\rightarrow$ Neuro}, where we take a symbolic domain (reductions of the $\lambda$ calculus) and apply it to a neural architecture (the Transformer).

\subsection{Neural Networks}

Artificial Neural Networks (NNs) were inspired by the structure and function of the human brain and are designed to process large amounts of data to identify patterns and relationships. Their fundamental unit is the neuron, which essentially "activates" when a linear combination of its inputs surpasses a certain threshold. A Neural Network is merely a collection of interconnected neurons whose properties are determined by the arrangement of the neurons and their characteristics \cite{Russell2021}.

These neurons are often organized into layers. The input data are fed into the first layer, and the output of each neuron in a given layer is used as the input for the next layer until the final layer produces the network output. The connections between the neurons are represented by weights that are updated during the training process to minimize the error between the predicted output and the actual output.

NNs have been applied to a wide range of tasks, including image classification, speech recognition, and natural language processing. One of the main advantages of NNs is their ability to model nonlinear relationships between inputs and outputs. This makes NNs a powerful tool for solving complex real-world problems. However, traditional NNs have fixed-size inputs and outputs, which are not suitable for our desired tasks, which have inputs and outputs of variable sizes.

\subsection{Sequence-to-sequence Models}

Although neural networks are versatile and effective, they are only suitable for problems where inputs and targets can be represented by fixed-dimensional vector encodings. This is a significant constraint, as many crucial problems are better expressed using sequences of unknown lengths, such as speech recognition and machine translation. It is evident that a versatile method that can learn to translate sequences to sequences without being restricted to a specific domain would be valuable \cite{Sutskever2014}.

Sequence-to-sequence (seq2seq) models emerged from this need. They are a type of deep learning model that is used for tasks that involve mapping an input sequence to an output sequence of variable length. They have traditionally been applied to various natural language processing tasks, such as machine translation, text summarization, and text generation. The assembly of seq2seq models can be done according to different model architectures such as RNN \cite{lipton2015critical}, LTSM \cite{Sutskever2014},  GRU \cite{chung2014empirical} and Transformer \cite{Vaswani2017}. 

Given that we can see the tasks we want to accomplish in this work as machine translation tasks, we have opted to employ the sequence-to-sequence model using the Transformer.



\subsection{The Transformer Model}

The Transformer model is a type of neural network architecture that was introduced in \cite{Vaswani2017}. It is designed to handle sequential data, such as natural language, and has quickly become one of the most popular models for tasks such as natural language processing, machine translation, text classification, and question answering.
One of the key innovations of the Transformer model is its use of a self-attention mechanism, which allows the model to dynamically weigh the importance of different parts of the input sequence. This allows the Transformer to capture long-range dependencies in the data, which is particularly useful for processing sequences of variable lengths.
Another advantage of the Transformer is its parallelization capacity, which allows it to be trained efficiently on large amounts of data. The Transformer model can be trained in parallel on multiple sequences, which is not possible with other traditional sequence-to-sequence models. 

 We chose the Transformer model mainly because to perform the $\beta$-reduction over lambda terms, it is necessary to substitute every occurrence of the variable in question with a term, and we believe that the self-attention mechanism can be used to ``pay attention'' to all these variable occurrences when performing the task.


\section{Building Experiments} \label{sec4}

For each of the hypotheses outlined in the Introduction, we propose a different task for our model to learn. The hypothesis \textbf{H1}  claims the model can perform a single-step computation in the $\lambda$-Calculus, that is, it can take a $\lambda$-term and transform it according to the single-step $\beta$-reduction rules of Definition \ref{def:one-step-rules} following a normal order strategy. We call this the One-Step Beta Reduction (OBR) task.

The hypothesis \textbf{H2} is that the model can perform multiple reduction steps in Lambda Calculus, taking a normalizable $\lambda$-term, i.e., a $\lambda$-term that has a normal form, and transforming it into its normal form through multiple beta reduction steps, following a normal order strategy. We call the task associated with this hypothesis the Multi-Step Beta Reduction (MBR) task. 

The primary focus of our research question aligns with the second hypothesis. However, we chose to begin with a more straightforward hypothesis as a starting point. The first task is considered easier because it requires the execution of a single computational step, which is less complex than performing a full computation. This approach helps us gradually build our understanding and confidence before moving on to the more challenging second hypothesis.

To support these hypotheses, we generate several datasets for each of the tasks and use them to train machine learning models. By training the models on these datasets, we will determine if the models can learn and perform the tasks associated with each hypothesis.

 \subsection{On Training}

To learn the tasks mentioned before, we use a neural model. Since the $\lambda$-terms we are using do not have a fixed size, we need our model to accept inputs of varying lengths and generate outputs accordingly. To achieve this, we use a sequence-to-sequence model (seq2seq), which allows for inputs and outputs of different sizes. Specifically, we use the Transformer model proposed by \cite{Vaswani2017}. This model has been widely used for various applications, including symbolic ones, as demonstrated by \cite{lample2019}.

For the hyperparameters, preliminary tests showed us that the parameters used by \cite{lample2019} were good enough for our tasks. If needed, they can be adjusted in the training process. So, the initial hyperparameters are the following:

\begin{itemize}
    \item Number of encoding layers - 6
    \item Number of decoding layers - 6
    \item Embedding layer dimension - 1024
    \item Number of attention heads - 8
    \item Optimizer - Adam \cite{kingma2014adam}
    \item Learning rate - $1 \times 10^{-4}$
    \item Loss function - Cross Entropy
\end{itemize}

\subsubsection{Experimental Setting} \label{sec:config}

The experiments were conducted on a server with the following configuration:

\begin{itemize}
    \item CPU: Intel(R) Core(TM) i7-8700 CPU @  3.20GHz
    \item RAM: 32 GB (2 x 16 Gb) DDR4 @  2667 MHz
    \item GPU: Quadro P6000 with 24 Gb
    \item OS: Ubuntu 18.04.5 LTS
\end{itemize}

Our initial goal was to run each training session for 12 to 24 hours. The preliminary results showed that each training consisting of $50$ epochs with an epoch size of $50000$ would take between 12 and 30 h to complete. So, we chose this arrangement. This configuration allows the model to process a total of $2.5 \times 10^{6}$ equations, which is $2.5$ times the size of the dataset.

With this machine, model, and configuration, we can safely have inputs with up to 250 tokens. With more than that, we end up with a memory shortage.

\subsection{Lambda Sets and Datasets}
\label{sec:ls-datasets}

All the datasets generated for this work are simple text files with each line in the format 
\[ 
\mathtt{BETA ~M~ N} 
\]
For the OBR task, the pair of terms in \texttt{BETA M N} is interpreted as: $\lambda$ term $N$ results of a one-step reduction of $\lambda$ term $M$ following the normal order strategy. For the MBR task, \texttt{BETA M N} is interpreted as: $\lambda$ term $N$ is the normal form that results from a multi-step reduction of $\lambda$ term $M$, following the normal order strategy.

To generate the datasets on which the models will train, we first generate \textit{Lambda Sets} (LSs) containing only lambda terms.  From these LSs, we generate the datasets needed for the training. We start generating the following three LSs:

\begin{itemize}
    \item \textbf{Random Lambda Set (RLS)}: This LS is generated as random and unbiased as possible. Thus, this LS can have terms that do not have a normal form, and also terms that are codifications of Boolean expressions.  With the datasets generated from this LS, we want to assert that the model can learn $\beta$-reduction, regardless of whether the input terms represent meaningful codifications or have a normal form. 
    \item \textbf{Closed Bool Lambda Set (CBLS)}: This LS has its terms representing closed Boolean expressions. Thus, all the terms in this LS have a normal form. With the datasets generated from this LS, we want to assert that the model can learn the $\beta$-reductions from meaningful codifications. In addition, these datasets are useful to validate the models trained with the RLS datasets, i.e., to validate whether the model learned from random terms can perform computations on terms that are meaningful codification.
    \item \textbf{Open Bool Lambda Set (OBLS)}: The terms in this LS are codifications of open Boolean expression, that is, with free variables in them. With the datasets generated from this LS, we want to assert that the model can learn $\beta$-reduction from terms that are meaningful codifications, even if they  have free variables.
\end{itemize}

For the One-Step Beta Reduction Task (OBR), we generate datasets based on the three LSs proposed. However, for the Multi-Step Beta Reduction task, we do not utilize the Random LS to generate datasets, as the terms produced randomly may not have a normal form and result in an infinite loop during the multi-step $\beta$-reduction. 

In addition to the LS mentioned above, we create an additional LS for each task, which we refer to as a \textbf{Mixed Lambda Set (MLS)}. For OBR, the MLS comprises terms coming from RLS, CBLS, and OBLS in the same proportion. For the MBR, the MLS is formed by terms from CBLS and OBLS, which are in the same proportion. With these mixed LSs, we want to assert that the model can learn from a domain that considers several kinds of terms.

From each of the four lambda sets presented above, we generate three datasets (all in the prefix notation), one for each of the following schemes for variable-naming:

\begin{itemize}
    \item \textbf{De Bruijn}: this variable naming convention, presented in Section ~\ref{sec:de_bruijn} is a way of representing $\lambda$-terms without naming the variables. It results in shorter terms and presents a factor that can be beneficial for our model.
    \item \textbf{Traditional}:  Datasets following this convention are generated from the datasets in the de Bruijn notation by replacing the de Bruijn indexes by traditional variable names, such as $a$, $b$, $c$, $x$, $y$, $z$, etc, following alphabetical order. We utilize this convention because we want to compare the results of the learning process using the de Bruijn convention  with those using the traditional convention
    \item \textbf{Random Vars}: This convention also uses the traditional convention for variable naming.  However, for this version, we take the de Bruijn datasets and replace the de Bruijn indexes by traditional variable names chosen randomly. We utilize this approach because we want to check if the way we name the variables matters for the models' accuracy.
\end{itemize}

\begin{figure}[H]
     \centering
         \includegraphics[width=\columnwidth]{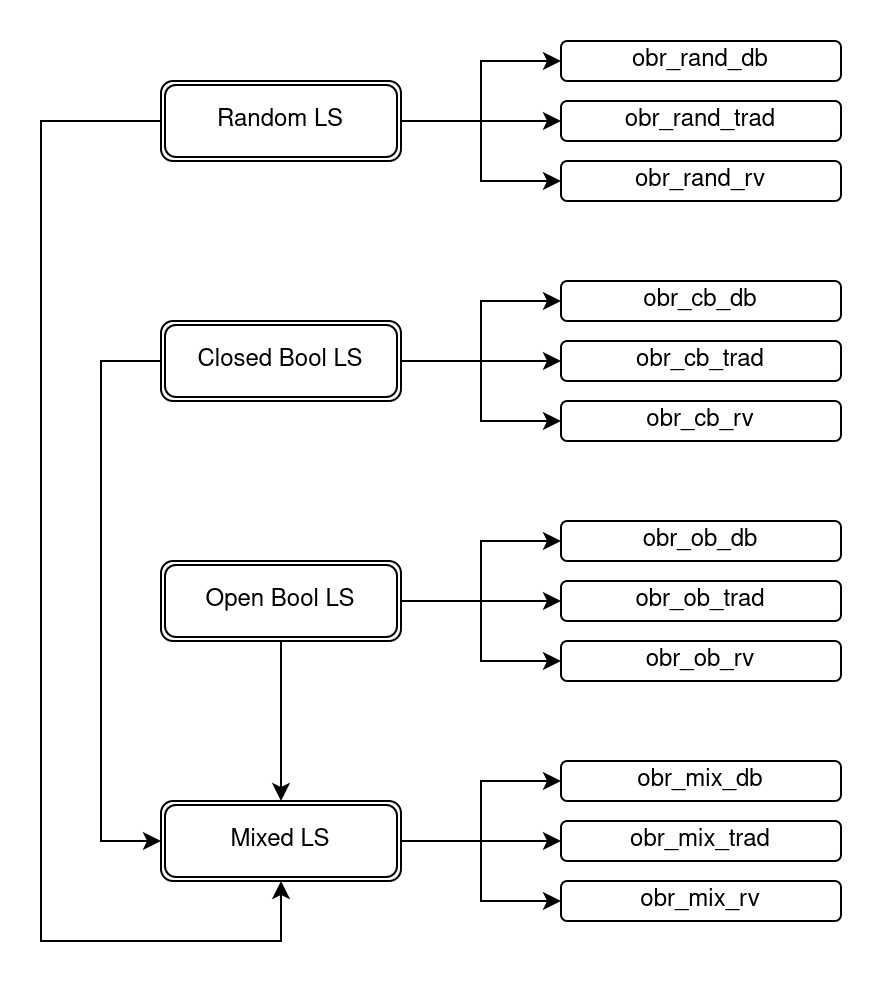}
         \caption{Scheme of how all the datasets for the OBR tasks are generated. It starts with the three Lambda Sets (RLS, CBLS, and OBLS) and ends with all 12 datasets available for the OBR task.}
         \label{fig:obr_datasets}
\end{figure}

\begin{figure}[H]
     \centering
         \includegraphics[width=\columnwidth]{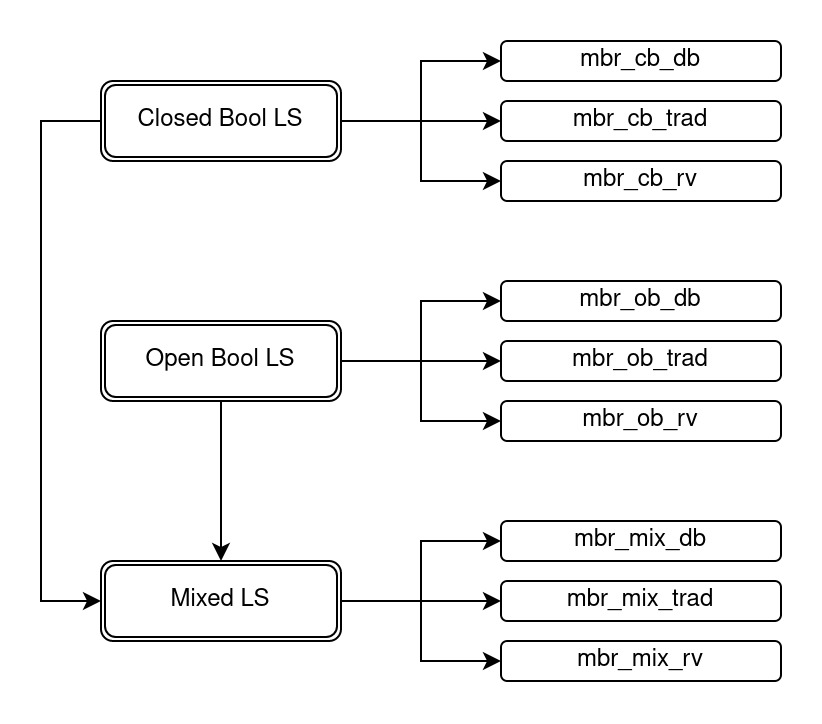}
         \caption{Scheme of how all the datasets for the MBR tasks are generated. It starts with the two Lambda Sets (CBLS, and OBLS), and ends with all nine datasets that are available for the MBR task.}
         \label{fig:mbr_datasets}
\end{figure}

In summary, for the OBR task, we ultimately have a total of 12 datasets, as illustrated in Figure ~\ref{fig:obr_datasets}, and for the MBR task, we have nine datasets, as shown in Figure ~\ref{fig:mbr_datasets}. It should be noted that the same CBLS and OBLS are utilized by both the OBR and the MBR tasks, meaning that the initial $\lambda$-terms for the datasets that employ these same LS are consistent across both tasks. These datasets provide us with a broad set of test cases to evaluate the performance of our models. 

Each dataset contains around one million examles (pairs of terms \texttt{BETA M N}), and we further divide each dataset into training, validation, and test sets. Keeping with the methodology described by \cite{lample2019}, we allocate approximately ten thousand examples for both the validation and the test sets. This division of the datasets into training, validation, and test sets allows us to effectively train our models, tune their parameters, and evaluate their performance on independent data. Using a large number of examples in each dataset and following established best practices, we want to ensure that our results are robust and representative of the underlying task.

\subsubsection{Term Sizes}

As mentioned in Section ~\ref{sec:config}, the maximum number of tokens that our configuration allows is 250. 
With this restriction, we were able to generate the terS sizes listed in Table ~\ref{tab:sizes}. Although we calculated these sizes using the datasets with terms in the traditional variable-naming convention, we expect similar results for the other datasets (with the exception of the de Bruijn convention, which should produce smaller term sizes).

\begin{table}[h]
\centering
\begin{tabular}{|l|l|c|c|c|}
\hline
Task                 & Dataset      & min & max & avg           \\ \hline
\multirow{4}{*}{OBR} & random      & $5$   & $249$ & $127.2 \pm 64.99$ \\ 
                     & closed bool & $9$   & $193$ & $97.6 \pm 26.76$  \\  
                     & open bool   & $5$   & $181$ & $66.46 \pm 21.73$ \\  
                     & mixed       & $5$   & $249$ & $86.93 \pm 46.56$ \\ \hline
\multirow{3}{*}{MBR} & closed bool & $9$   & $193$ & $97.55 \pm 26.75$ \\  
                     & open bool   & $5$   & $181$ & $66.46 \pm 21.72$ \\ 
                     & mixed       & $5$   & $181$ & $77.96 \pm 28.02$ \\ \hline
\end{tabular}
\caption{Table showing the minimum, maximum, and average sizes of the input $\lambda$-terms for each dataset. The datasets  considered were the ones that use traditional variable naming.}
\label{tab:sizes}
\end{table}

\subsubsection{Number of Reductions}

For certain Lambda sets, we iteratively generate the reductions until a normal form is reached for each term. Thus, another important metric we consider is the number of reductions each term had to undergo to reach its normal form. This number can be seen as the number of computational steps necessary to evaluate a given term. Table ~\ref{tab:reduction_sizes} shows the average number of reductions that the terms of each Lambda Set have undergone to generate their respective datasets for the OBR and MBR tasks.

\begin{table}[h]
\centering
\begin{tabular}{|l|c|c|c|}
\hline
Lambda Set  & min & max & avg           \\ \hline
closed bool & $3$   & $100$ & $18.8 \pm 12.22$  \\ \hline
open bool   & $1$   & $100$ & $18.88 \pm 10.42$ \\ \hline
mixed       & $2$   & $100$ & $18.82 \pm 11.32$ \\ \hline
\end{tabular}
\caption{Table showing the minimum, maximum, and average number of reductions that the terms of each Lambda Set have undergone  The Mixed Lambda Set considered here is the one with terms coming only from the closed bool and open bool Lambda Sets.}
\label{tab:reduction_sizes}
\end{table}

\subsection{Code and Implementation}

In this work, we used two distinct pieces of code. The first piece of code, adapted from \cite{lample2019}, was responsible for generating what we call \textit{intermediate}  $\lambda$-terms. These intermediate lambda terms follow the syntax given by the grammar in Definition \ref{def:grammar}. This piece of code also handled the learning process.

The second piece of code is responsible for generating, from the intermediate lambda terms, the lambda terms in the formats required for the lambda sets, as described in Section ~\ref{sec:ls-datasets}. It also implements a simulator that performs reductions of these lambda terms. From the simulation results, it generates the datasets with pairs $\mathtt{BETA ~M~ N}$ where the lambda term $N$ is the result of the lambda term $M$ reduction. 

 We observe that after generating the datasets, they must first be cleaned by (i) deleting repeated pairs, and (ii) because we want all pairs to represent $\beta$-reductions, we also delete all pairs $\mathtt{BETA ~M~ N}$ with the first component $M$ in normal form.

\subsection{Accuracy and Similarity} \label{sec:meth_experiments}

To evaluate the performance of the model, its accuracy in predicting the data in the test data set is calculated and recorded after each epoch. This accuracy metric measures how well the model can make correct predictions on the data it has seen during training. For each of the models trained, we display a graph showing the evolution of the model's accuracy (y-axis) over the epochs (x-axis).

The accuracy of the model determines whether the predicted string matches the expected output. However, accuracy may not be the only relevant metric for evaluating the performance of a model in text generation or other similar tasks. In some cases, it may be useful to measure the similarity between the predicted and expected strings, even if they are not identical.
So, additionally, we used a string similarity metric to compare how close the predicted string is to the expected one. For this, we used a common string similarity metric, the Levenshtein distance, which measures the number of changes (insertions, deletions, or substitutions) needed to transform one string into the other \cite{levenshtein1966binary}. This metric provides the absolute difference between the two strings, so we divide this distance by the maximum distance possible between the two strings (which is the length of the longer string) to generate a percentage of dissimilarity. Then, we subtract this value from 1 to get a percentage of similarity between the two strings. So, we also provide a string similarity value for each trained model. The formula used is as follows:
\[
\mathsf{str\_sim}\, (s_1, s_2) = ~~ 1 - \frac{\mathsf{lev\_dist}\,(s_1,s_2)}{\mathsf{max}\, (\mathsf{len}(s_1), \mathsf{len}(s_2))}
\]
As part of our analysis, we also assessed the capacity of the models trained with each dataset to evaluate the other datasets. We achieve this by performing additional evaluations with each of the already trained models. For this, we use a model trained with one dataset to evaluate the other datasets that use the same notation and are designed for the same task. For example, we take the model that trained on the \textit{obr\_rand\_trad} dataset and evaluate how it performs on the \textit{obr\_cb\_trad}, \textit{obr\_ob\_trad} and \textit{obr\_mix\_trad} datasets with respect to accuracy and similarity.

By evaluating a model on the other datasets, which it did not train on, we can better understand its strengths and weaknesses and its ability to generalize to new data. If the model performs well on other datasets of the same type, it may be a good sign that it has learned meaningful patterns in the data and can be applied to new, unseen data. If the model performs poorly on other datasets, it may indicate that the model has, for instance, overfit to the original training data or the data was not adequate.


\section{Experimental Results} \label{sec5}

Some trainings experienced an oscillation in accuracy, indicating that the initial learning rate ($1 \times 10^{-4}$) was too high. So, we had to adjust the learning rate for these trainings. We initially used the same value for all trainings, but decreased it based on the degree of accuracy oscillation. Table ~\ref{tab:learning-rates} shows the final learning rates for each training performed. Although the learning rate was adjusted for different trainings, we kept it consistent for the three conventions in each dataset for comparison purposes. 
It is important to note that we did not thoroughly search for the optimal learning rate; instead, we selected a parameter that resulted in satisfactory and converging accuracy.
\begin{table}[h]
\centering
\begin{tabular}{|c|l|l|}
\hline
Task & Lambda Set & Learning Rate \\ \hline
\multirow{4}{*}
{\begin{tabular}[c]{@{}l@{}}
One-Step Beta \\ 
Reduction
\end{tabular}} 
& random        & $1 \times 10^{-4}$             \\ 
& closed bool   & $6 \times 10^{-5}$              \\ 
& open bool     & $8 \times 10^{-5}$              \\  
& mixed        & $1 \times 10^{-4}$              \\ 
\hline
\multirow{3}{*}
{\begin{tabular}[c]{@{}l@{}}
Multi-Step Beta \\ 
Reduction
\end{tabular}}  
& closed bool   & $3 \times 10^{-5}$              \\  
& open bool     & $5 \times 10^{-5}$              \\
& mixed        & $5 \times 10^{-5}$               \\ 
\hline
\end{tabular}
\caption{Values for the learning rate hyperparameter chosen for each of the tasks and lambda sets trained. The value started with $1 \times 10^{-4}$, and it was lowered as the trained showed an unacceptable oscillation, indicating the learning would not converge.}
\label{tab:learning-rates}
\end{table}

\subsection{Training Results}

This section presents graphs that illustrate the training results for every model trained.  The graphs display the model's accuracy for the test dataset of the respective training dataset as it evolves over the training epochs. 
Each graph presents the results for all three variable naming conventions utilized in this study: the traditional convention, the random vars convention, and the de Bruijn convention.

\begin{figure}[h]
     \centering
         \includegraphics[width=\columnwidth]{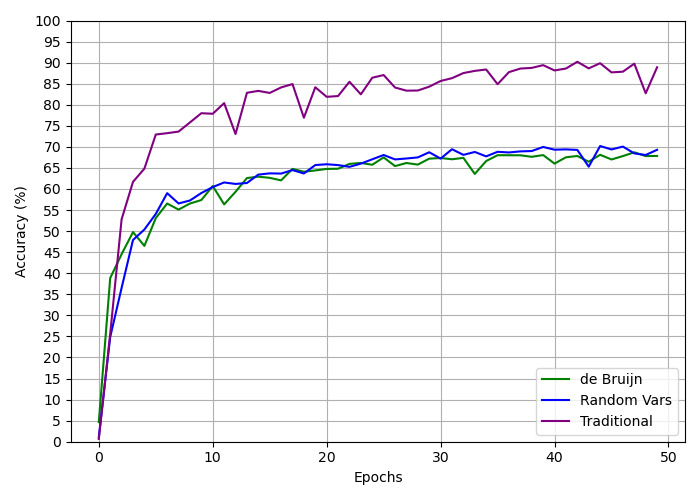}
         \caption{Graph displaying the progression for the training of the One-Step Beta Reduction task, for the random dataset, over the three different conventions.}
         \label{fig:obr-random}
\end{figure}

\begin{figure}[h]
     \centering
         \includegraphics[width=\columnwidth]{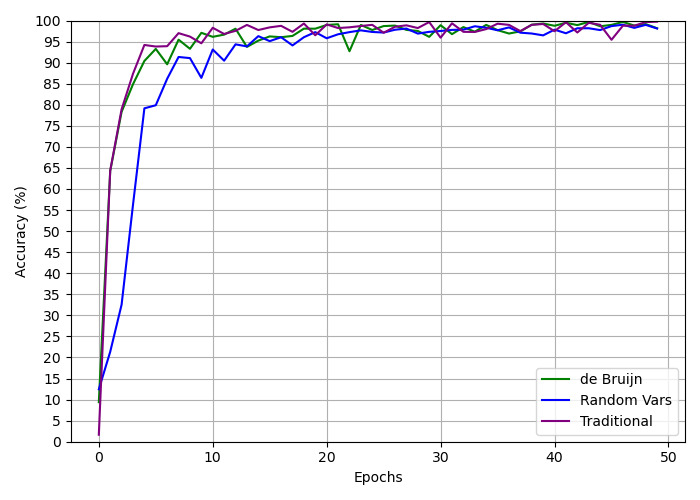}
         \caption{Graph displaying the progression for the training of the One-Step Beta Reduction task, for the closed bool dataset, over the three different conventions.}
         \label{fig:obr-cb}
\end{figure}

\begin{figure}[h]
     \centering
         \includegraphics[width=\columnwidth]{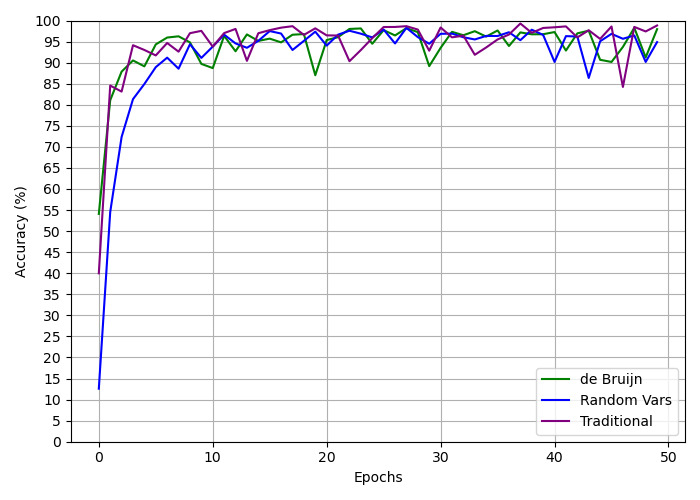}
         \caption{Graph displaying the progression for the training of the One-Step Beta Reduction task, for the open bool dataset, over the three different conventions.}
         \label{fig:obr-ob}
\end{figure}

\begin{figure}[h]
     \centering
         \includegraphics[width=\columnwidth]{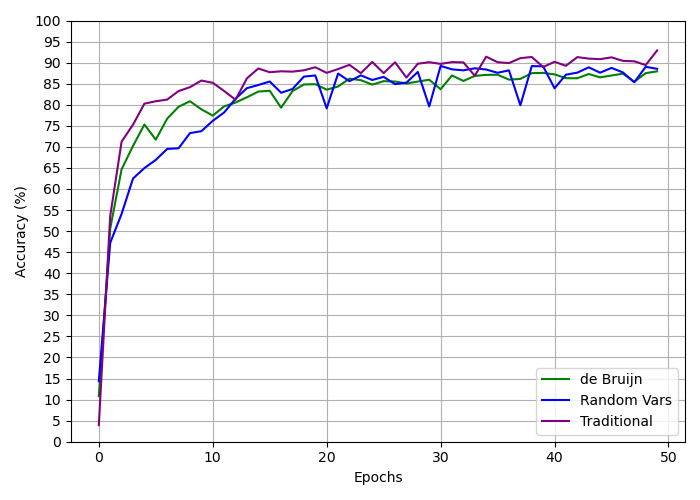}
         \caption{Graph displaying the progression for the training of the One-Step Beta Reduction task, for the mixed dataset, over the three different conventions.}
         \label{fig:obr-mix}
\end{figure}

\begin{figure}[h]
     \centering
         \includegraphics[width=\columnwidth]{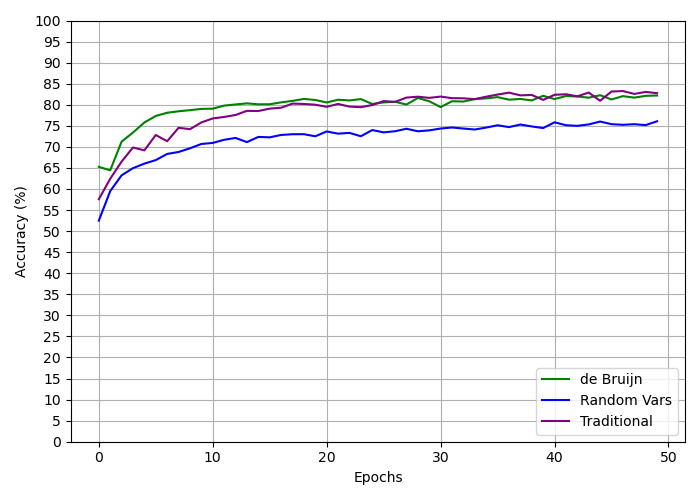}
         \caption{Graph displaying the progression for the training of the Multi-Step Beta Reduction task, for the closed bool dataset, over the three different conventions.}
         \label{fig:mbr-cb}
\end{figure}

\begin{figure}[h]
     \centering
         \includegraphics[width=\columnwidth]{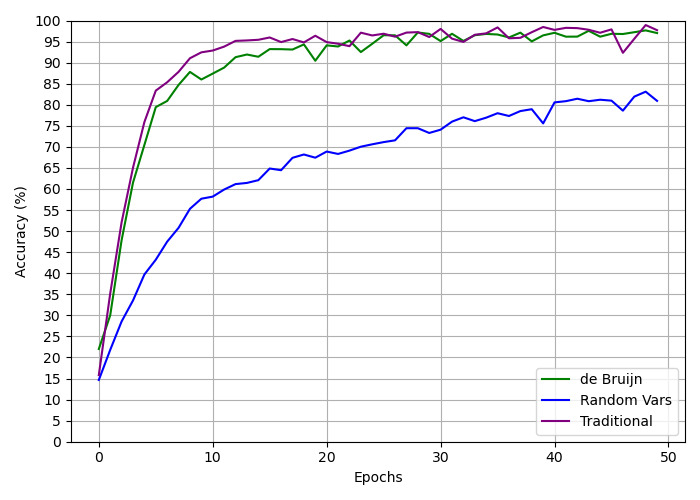}
         \caption{Graph displaying the progression for the training of the Multi-Step Beta Reduction task, for the open bool dataset, over the three different conventions.}
         \label{fig:mbr-ob}
\end{figure}

\begin{figure}[h]
     \centering
         \includegraphics[width=\columnwidth]{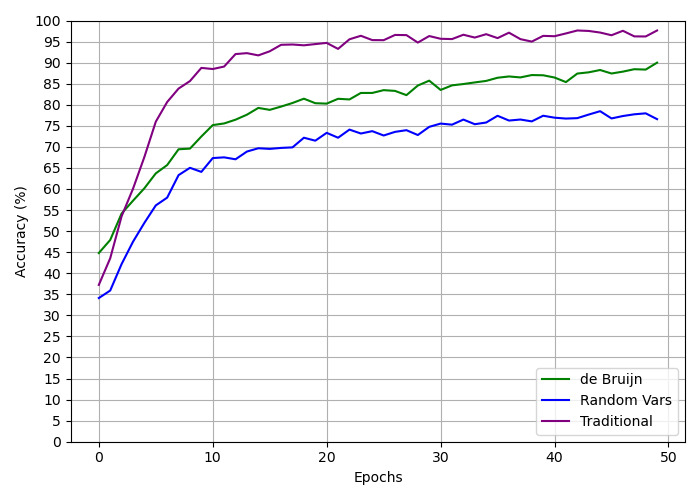}
         \caption{Graph displaying the progression for the training of the Multi-Step Beta Reduction task, for the mixed dataset, over the three different conventions.}
         \label{fig:mbr-mix}
\end{figure}

For the OBR task, the training for the random datasets can be seen in Figure ~\ref{fig:obr-random}, the closed bool datasets in Figure ~\ref{fig:obr-cb}, the open bool datasets in Figure ~\ref{fig:obr-ob}, and the mixed datasets in Figure ~\ref{fig:obr-mix}. 
For the MBR task, the training for the closed bool datasets can be seen in Figure ~\ref{fig:mbr-cb}, the open bool datasets in Figure ~\ref{fig:mbr-ob}, and the mix datasets in Figure ~\ref{fig:mbr-mix}.

In addition to the graphs, tables ~\ref{tab:accuracy-general-obr} and ~\ref{tab:accuracy-general-mbr} show the final accuracies, i.e. the accuracy of the last epoch for all the models trained for both the OBR and MBR tasks. The table also shows the average percentage of similarity of the strings, calculated using the Levenshtein distance shown in Section ~\ref{sec:meth_experiments}.
\begin{table*}[h]
\centering
\begin{tabular}{|l|l|c|c|}
\hline
 \multicolumn{1}{|c|}{Lambda Set} & \multicolumn{1}{c|}{Convention} & ACC (\%) & STR SIM (\%) \\ \hline
\multirow{3}{*}{random}         & trad                          & $88.89$    & $99.83$  \\ 
                                                             & random vars                  & $69.30$    & $99.51$  \\ 
                                                             & de Bruijn                    & $67.84$    & $99.34$  \\  \hline
                             \multirow{3}{*}{closed bool}    & trad                          & $99.73$    & $100.00$ * \\ 
                                                             & random vars                  & $98.10$    & $99.98$  \\ 
                                                             & de Bruijn                    & $98.16$    & $99.99$  \\ \hline 
                             \multirow{3}{*}{open bool}      & trad                          & $98.82$    & $99.99$  \\ 
                                                             & random vars                  & $94.88$    & $99.95$  \\ 
                                                             & de Bruijn                    & $97.94$    & $99.97$  \\ \hline
                             \multirow{3}{*}{mixed}            & trad                          & $92.88$    & $99.89$  \\ 
                                                             & random vars                  & $88.52$    & $99.77$  \\ 
                                                             & de Bruijn                   & $87.93$    & $99.73$  \\ \hline
\end{tabular}
\caption{Accuracy and the average string similarity for the evaluation of the models trained for the OBR task. * Rounded from $0.998$.}
\label{tab:accuracy-general-obr}
\end{table*}
\begin{table*}[ht]
\centering
\begin{tabular}{|l|l|l|c|c|}
\hline
\multicolumn{1}{|c|}{Lambda Set} & \multicolumn{1}{c|}{Convention} & ACC (\%) & STR SIM (\%) \\ \hline
\multirow{3}{*}{closed bool}    & trad                          & $82.75$    & $97.97$  \\  
                                                            & random vars                  & $76.08$    & $97.06$  \\ 
                                                             & de Bruijn                    & $82.20$    & $96.49$  \\ \hline
                             \multirow{3}{*}{open bool}      & trad                          & $97.70$    & $99.92$  \\  
                                                             & random vars                  & $80.92$    & $98.19$  \\ 
                                                             & de Bruijn                     & $97.02$    & $99.77$  \\ \hline
                             \multirow{3}{*}{mixed}            & trad                          & $97.63$    & $99.89$  \\ 
                                                             & random vars                  & $76.58$    & $98.15$  \\ 
                                                             & de Bruijn                   & $89.99$    & $98.64$  \\ \hline

\end{tabular}
\caption{Accuracy and the average string similarity for the evaluation of the models trained for the MBR task.}
\label{tab:accuracy-general-mbr}
\end{table*}

\subsection{Evaluation Across Datasets}    
\label{sec:eval_across_data}

In this section, we show the results obtained by some additional evaluations with the already trained models for the OBR and MBR tasks. We use a model trained with one dataset to evaluate the other datasets that use the same convention and are designed for the same task. For example, we take the model that trained on the \textit{obr\_cb\_db} dataset and evaluate how it performs on the \textit{obr\_rand\_db}, \textit{obr\_ob\_db} and \textit{obr\_mix\_db} datasets. We use test datasets to perform these evaluations.

Tables ~\ref{tab:accuracy_eval_obr} and ~\ref{tab:accuracy_eval_mbr} show the values found for the evaluations with the trained models for the OBR and MBR tasks, respectively.
\begin{table*}[h]
\centering
\begin{tabular}{|l|l|c|c|c|c|c|}
\hline
 Convention                                                             &  Lambda Set        & random & closed bool & open bool & mixed   & AVERAGE \\ \hline
\multirow{4}{*}{traditional}                                                  & random   & 88.89  & 63.69    & 72.66  & 80.47 & 76.43   \\ 
                                                                       & closed bool & 0.00   & 99.73    & 7.73   & 35.92 & 35.85   \\  
                                                                       & open bool   & 0.04   & 80.01    & 98.82  & 63.79 & 60.67   \\ 
                                                                       & mixed      & 72.26  & 97.42    & 99.62  & 92.88 & 90.55   \\ \hline
\multirow{4}{*}{\begin{tabular}[c]{@{}l@{}}random\\ vars\end{tabular}} & random   & 69.30  & 22.17    & 42.27  & 64.65 & 49.60   \\ 
                                                                       & closed bool & 0.05   & 98.10    & 18.82  & 39.26 & 39.06   \\  
                                                                       & open bool   & 0.17   & 77.24    & 94.88  & 61.53 & 58.46   \\ 
                                                                       & mixed      & 65.77  & 83.90    & 85.92  & 88.52 & 81.03   \\ \hline
\multirow{4}{*}{de Bruijn}                                             & random   & 67.84  & 47.15    & 58.35  & 61.14 & 58.62   \\ 
                                                                       & closed bool & 0.00   & 98.16    & 10.96  & 36.87 & 36.50   \\ 
                                                                       & open bool   & 0.01   & 77.93    & 97.94  & 58.59 & 58.62   \\ 
                                                                       & mixed      & 65.70  & 96.39    & 98.71  & 87.93 & 87.18   \\ \hline
\end{tabular}
\caption{Accuracy (\%) for the evaluation of the models over different datasets, for the task of one-step beta reduction. For each of the three different conventions (trad, random vars, and De Bruijn), the model trained with each dataset (rows) was evaluated with each dataset (columns). The last column indicates the average accuracy of the model over the different datasets.}
\label{tab:accuracy_eval_obr}
\end{table*}
\begin{table*}[h]
\centering
\begin{tabular}{|l|l|c|c|c|c|}
\hline
 Convention                                                             &  Lambda Set        & closed bool & open bool & mixed   & AVERAGE \\ \hline
\multirow{3}{*}{traditional}                                                  & closed bool & 82.75    & 15.85  & 49.66 & 49.42   \\ 
                                                                       & open bool   & 92.21    & 97.70  & 96.86 & 95.59   \\
                                                                       & mixed      & 96.20    & 93.28  & 97.63 & 95.70   \\ \hline
\multirow{3}{*}{\begin{tabular}[c]{@{}l@{}}random\\ vars\end{tabular}} & closed bool & 76.08    & 24.79  & 50.98 & 50.62   \\ 
                                                                       & open bool   & 75.23    & 80.92  & 84.25 & 80.13   \\  
                                                                       & mixed      & 72.72    & 50.68  & 76.58 & 66.66   \\ \hline
\multirow{3}{*}{de Bruijn}                                                    & closed bool & 82.20    & 20.20  & 57.83 & 53.41   \\  
                                                                       & open bool   & 90.02    & 97.02  & 95.37 & 94.14   \\ 
                                                                       & mixed      & 88.43    & 78.64  & 89.99 & 85.69   \\ \hline
\end{tabular}
\caption{Accuracy (\%) for the evaluation of the models over different datasets, for the task of multi-step beta reduction. For each of the three different conventions (trad, random vars, and De Bruijn), the model trained with each dataset (rows) was evaluated with each dataset (columns). The last column indicates the average accuracy of the model over the different datasets.}
\label{tab:accuracy_eval_mbr}
\end{table*}

\section{Discussion} \label{sec6}

In this section, the results obtained from the experiments carried out in the study are discussed in detail. The results are analyzed and interpreted to draw conclusions about the objectives of the study.

\subsection{Training}

In this section, we discuss the results of the trainings presented in Section ~\ref{sec4}. With these results, we are able to determine which datasets and conventions performed better and try to conjecture some hypotheses about what happened in the trainings.

\subsubsection{One-Step Beta Reduction}

In the training of the models for the OBR task, it was found that each model achieved an accuracy of at least $67.84\%$. However, when only the best conventions were considered, each model achieved a minimum accuracy of $88.89\%$. Furthermore, a model achieved remarkable accuracy $99.73\%$. These findings, which can be seen in Table ~\ref{tab:accuracy-general-obr},  highlight the high level of accuracy and effectiveness of the models, particularly when utilizing optimal conventions, which supports hypothesis H1.

The similarity metric for the strings can also be seen in Table ~\ref{tab:accuracy-general-obr} and indicates that, despite incorrectly predicting some terms, the model was able to accurately predict a significant portion of those terms, with all similarities being at least $99.34\%$. If we take only the best conventions, this number goes up to $99.83\%$. In addition, some models achieved an outstanding performance of more than $99.99\%$ for this metric. For example, the model trained with the random dataset with the de Bruijn convention got a final accuracy of $67.84\%$. However, the string similarity metric for the same training was $99.34\%$. This illustrates how much closer to the correct answers were the incorrect answers obtained by the model.

Upon analyzing the performance of the models on different datasets, it is evident that the closed bool and open bool datasets were easier to learn compared to other datasets, as we can see in Figures ~\ref{fig:obr-random}, ~\ref{fig:obr-cb}, ~\ref{fig:obr-ob} and ~\ref{fig:obr-mix}. The Boolean datasets achieved good accuracies in a shorter span of time and presented similar results between themselves. The random dataset was the hardest to learn, as shown in Fgurie ~\ref{fig:obr-random}. We think this is due to the absence of more defined patterns among the terms. The mixed dataset, as expected, fell between random and boolean data sets in terms of difficulty to learn. However, its accuracy surprised us, since it was the most diverse dataset, which means that it learned to perform the OBR task both for random and Boolean terms, with high accuracies ($92.88\%$ for the optimal convention, as shown in Table ~\ref{tab:accuracy-general-obr}). 

It should be noted that, although the accuracies for the random and mixed data sets were comparatively lower than those of the Boolean datasets, the graphs illustrating their performance present a growing pattern, as illustrated in Figures ~\ref{fig:obr-random} and ~\ref{fig:obr-mix}. This indicates that further training with more epochs could yield higher accuracies for these datasets.

Analyzing the performance of the models that use different conventions in Table ~\ref{tab:accuracy-general-obr}, it can be observed that the traditional convention consistently outperformed the other two conventions, which exhibited similar levels of performance. However, the only training that the convention really made a difference was in the random dataset, which we saw was the hardest to learn. We suppose that for the other datasets, the difference in convention did not matter because it was so easy for the model to learn that even the ``harder'' conventions were not a problem.

We also think that the traditional convention performed overall better than the other two conventions because for the de Bruijn convention, despite being based on a simpler notation, the beta reduction is more intricate, and consequently, harder to learn. Furthermore, compared to the random-vars convention, the traditional convention, with its ordered naming rule, tends to provide the model with more predictable outcomes.

\subsubsection{Multi-Step Beta Reduction} \label{sec:msbr}

For the training of the models for the MBR task, every model achieved a minimum accuracy of $76.08\%$. However, when considering only the best conventions, each model exhibited an accuracy of at least $82.75\%$. Furthermore, one model achieved an exceptional $97.70\%$ accuracy. These results, which can be seen in Table ~\ref{tab:accuracy-general-mbr}, emphasize the effectiveness and high accuracy of the models, especially when using the optimal conventions, which supports hypothesis H2.

The similarity metric for the strings, found in Table ~\ref{tab:accuracy-general-mbr}, again indicates that, even though the model made incorrect predictions for some terms, it accurately predicted a significant portion of those terms, with all similarities no less than $96.49\%$. Considering only the best conventions, this number increases to $97.97\%$. Furthermore, some models performed exceptionally well, achieving up to $99.92\%$ for this metric. Again, the models that did not obtain a good accuracy exhibited outstanding performance in this metric. For example, the model trained with the mixed dataset, using the random vars convention, obtained an accuracy of $76.58\%$. However, the string similarity metric for the same training was $98.15\%$. This shows that for this task, the models also got the wrong predictions very close to the correct ones.

For the closed bool dataset in the MBR task, it is important to note that the set of possible terms that the model should predict is small (namely, \textit{true} and \textit{false}). For the traditional convention and, especially, for the random vars convention, the \textit{true} and \textit{false} terms are not always the same term, since there are many alpha-equivalent terms for \textit{true} and \textit{false} using the English alphabet. But in the DB case, there are only 2 distinct terms for the \textit{true} and \textit{false} (``\textit{L L 2}'' and ``\textit{L L 1}'', respectively). Thus, one might expect that the closed bool dataset would be easier to learn since there are only a few possible terms for the model to predict (only two in the DB case), while the open bool, on the other hand, had output terms that differ dramatically from one another. 

However, the opposite was actually observed, as we can see in Figures ~\ref{fig:mbr-cb} and ~\ref{fig:mbr-ob}. The closed bool dataset was found to be harder to learn than the open bool dataset, with the model that trained on it having significantly lower accuracy than the open bool model, which seems counter-intuitive. Our hypothesis is that, precisely because the terms were so similar in the closed bool dataset, the model resorted to guessing the output term from a limited set of possibilities, based on some features of the inputs, instead of learning to perform the reductions. But, since this was not possible for the open bool dataset, the model was forced to actually learn to perform the multi-step beta reduction on the input terms. The fact that the closed bool model already starts the training with around $55\%$ accuracy also corroborates to our hypothesis that the model is learning to guess from a limited set instead of learning the reductions. 

The model trained with the mixed dataset seems to have overcome this issue, as we can see in Figure ~\ref{fig:mbr-mix}. Considering the traditional convention, the accuracy of the model was similar to the accuracy of the model trained on the open bool dataset, even with half of its terms having come from the closed bool dataset. This actually supports our previous hypothesis, since we think that having more variability on the terms forced the model to learn the reductions instead of only guessing between a small set of possible outcomes.

For the trainings on different conventions, Table ~\ref{tab:accuracy-general-mbr} shows that the random vars convention had the worst accuracies for the three datasets. However, only the models trained on the open bool and on the mixed datasets presented a large gap between different conventions. We suppose that the naming convention did not change the guessing factor on the learning process for the models that trained on the closed bool. What is interesting is that the de Bruijn convention led to accuracies as good as the traditional convention and significantly better than the random vars convention for the models trained on the open bool and closed bool datasets. This was unexpected since the $\beta$ reduction in the de Bruijn notation is more intricate than in the traditional notation, which the other two conventions use. For the model trained on the mixed dataset, the order of the different conventions was more aligned with the expected, with the traditional being the best convention, followed by the other two. However, this result, although expected, was unusual, since the other models did not follow this order.

\subsection{Evaluations Across Datasets}

In this section, we discuss the results of the evaluations in the datasets presented in Section ~\ref{sec:eval_across_data}. These evaluations can give us some insight into whether the model really learned the reductions or whether it just learned the reductions for that specific set of terms. 

\subsubsection{One-Step Beta Reduction}

Evaluations of this task have yielded promising results, especially for models trained with the mixed dataset. It produced better average accuracies for all models in the OBR tasks, as seen in Table ~\ref{tab:accuracy_eval_obr}. This shows that, as expected, these models were able to better capture the diversity of terms present in the different datasets.

The models trained with both Boolean datasets performed poorly for the evaluation with the random dataset, with accuracies close to $0\%$, as we can see in Table ~\ref{tab:accuracy_eval_obr}. We suppose that this happened because the terms in the random data set are very distinct from the terms in the Boolean datasets. Also, since the opposite did not happen, we think that the random dataset contains terms that are actually harder to learn, as we presumed in the previous section. 

As expected, almost every model had better accuracy in the evaluation with the dataset in which it was trained than with the others, as shown in Table ~\ref{tab:accuracy_eval_obr}. But one result that may be seen as counterintuitive is the evaluation of the model trained with the mixed dataset. It had better accuracy for the open bool and closed bool datasets for the traditional and de Bruijn conventions. We again suppose that this happened because the random dataset is the hardest to learn. Thus, since the mixed dataset has 1/3 of its terms from the random dataset, it ends up being harder than the closed bool and open bool datasets. So, it ends up evaluating those two datasets better than the dataset it was trained with, which contains terms from the random dataset. 

Another interesting result is that the model trained on the open bool dataset was able to extrapolate and obtain good accuracies for the evaluation of the closed bool dataset, with a minimum accuracy of $77.24\%$. But the opposite did not happen, with accuracies as low as $7.73\%$, as we can see in Table ~\ref{tab:accuracy_eval_obr}.

Aside from what was mentioned, the different conventions did not present a significant difference between the evaluations.

\subsubsection{Multi-Step Beta Reduction}

The evaluations for this task have again yielded good results, particularly for the models trained with the open bool dataset. As shown in Table ~\ref{tab:accuracy_eval_mbr}, the use of the open bool dataset led to better average accuracies for almost all models in the MBR task.

Table ~\ref{tab:accuracy_eval_mbr} shows that, as expected, the majority of models performed better in the evaluations using the dataset they were trained on, as opposed to the other datasets. However, the models trained with the open bool dataset had accuracies quite close to one another for the three datasets evaluated. In fact, it presented better accuracy for the mixed dataset rather than on the dataset on which it was trained. We presume that this happened because the model trained with the open bool dataset was able to generalize better than the others.

Again, the models trained with the closed bool did not extrapolate and obtained good accuracies for the other datasets, especially the closed bool, with accuracies as low as $15.85\%$, as seen in Table ~\ref{tab:accuracy_eval_mbr}. But the opposite happened, with the open bool models getting a minimum of $75.23\%$ of accuracy for the closed bool dataset. We think that this happened for the same reason discussed in Section ~\ref{sec:msbr}, which is that the model trained on the closed bool dataset just learned to guess the output from a limited set of possible terms, not actually learning the $\beta$-reduction.
Apart from what was stated, the different conventions did not present any main differences between the evaluations.

\section{Conclusions and Further Work}

Through comprehensive experimentation and analysis, it was demonstrated that the Transformer model is capable of capturing the syntactic and semantic features of $\lambda$-calculus, allowing for accurate and efficient predictions. The results obtained were positive, with overall good accuracies for both tasks at hand. For the One-Step Beta Reduction, we got accuracies up to $99.73$\%, and string similarity metric of over $99.99$\%. For the Multi-Step Beta Reduction, we obtained accuracies of up to $97.70$\%, and string similarity metric exceeding $99.90$\%. In addition to that, the models presented a good generalization performance across different datasets. 

Due to limitations of hardware and time, our models trained for just 50 epochs. Considering that is a pretty low number compared with substantial trainings of large models and that we did not do a search for the optimal hyperparameters, we can assure that the accuracy of our models can be even higher than what was presented here.  Addressing the technological limitations of our work would provide a way for further advancements.

These results illustrate the effectiveness of the model in learning the desired tasks and support the two hypotheses raised in this study and, subsequently, the proposed research question. In addition, these results showed that the Transformer self-attention mechanism is well suited for capturing the dependencies between variables and functions in the $\lambda$-Calculus. 

We believe that the methods and results presented in this work have yielded some significant results for future research. The main contributions that have resulted from this research can be summarized as follows.

\begin{itemize}
\item  Lambda calculus learning: The outcomes from learning the reductions of Lambda Calculus are promising and hold potential implications for future research in the field of AI and computer programs.

    \item Dataset generation: Since datasets for lambda terms and reductions did not exist, we implemented the generation for these datasets from scratch. These datasets and generation methods can be used in future research in the Lambda Calculus domain.
    
    \item Functional programming learning: The results obtained in this study can be taken into account to shift the programming paradigm from imperative to functional in future research in the field of learning to compute.
\end{itemize}%
 Of course, our contributions are mainly in the proposal of a novel neural Lambda Calculus, and the experiments we carried out serve to illustrate the potential of our approach in AI and programming languages. 
Furthermore, although our research was limited to one formalism, the $\lambda$-calculus, this opens the possibility of further analyzes of other computational calculi. Thus, with respect to future research, we propose the following.\\
\noindent 1. Further training experiments: The current study trained the model for a limited number of epochs. Further research could aim to train the best notation for more epochs to see if performance can be improved.\\ 
\noindent 2. Hyperparameter optimization: The study used a set of predefined hyperparameters for the transformer model. A thorough search for the optimal hyperparameters could be conducted to find the best set of hyperparameters for learning Lambda Calculus. \\
\noindent 3. Improved error analysis: The study provided a preliminary error analysis, but more work could aim to conduct a more in-depth error analysis to better understand the types of mistakes the model is making and to identify areas for improvement. \\
\noindent 4. Incorporating other formalisms: This study focused on learning Lambda Calculus, but there are other formalisms such as Combinatorial Logic and Turing Machines that could be trained by the model and compared with the current work. \\
\noindent 5. Solve typing problems: Learn how to solve some typing problems (well-typedness, type assignment, type checking, and type inhabitation \cite{pierce2002types}) that can be uncomputable for some typed $\lambda$-Calculus   \\
\noindent 6. Learn more complex versions of the $\lambda$-Calculus: Learn the non-pure $\lambda$-Calculus, with numbers and arithmetical and Boolean operations already embedded. \\
\noindent 7. Learn to compute a functional programming language: Learn a functional programming language based on $\lambda$-Calculus, such as Haskell or Lisp \cite{thompson2011haskell}. \\
\noindent 8. Learn to detect loops: Use the same methods for training, but instead of learning to perform the computations, learn to identify if a $\lambda$-term does not have a normal form, i.e., if it is going to enter a loop when applying the reductions.\\ 
\noindent 9. The Curry-Howard
isomorphism \cite{barendregt1997impact} establishes a relationship between intuitionistic logic and the typed $\lambda$-calculus. As already advocated by \cite{GarcezLG06}, it would be relevant to further exploit this relationship with respect to the connectionist model presented in \cite{GarcezLG06,Garcez2009}. 

These future work suggestions have the potential to bring further advancements in the application of machine learning models to the field of symbolic learning. In particular, with respect to programming languages in general and in particular regarding the use of functional programming as the base paradigm. In summary, we believe that a neurosymbolic approach in which a neural lamda calculus is a foundation can contribute to a deeper understanding of the underlying computational processes in AI.


\section*{Declarations}
L.C. Lamb and J.M. Flach contributed for this work in the conceptualization of the study. J.M. Flach performed the experiments. A.F. Moreira contributed to the aspects of Lambda Calculus. All authors contributed to the writing and revision. All authors read and approved the final manuscript. 
The authors declare that they have no competing interests. 
 This research was supported in part by CAPES Finance Code 001 and the Brazilian Research Council CNPq. 
 The datasets generated during the current study are available at \url{https://bit.ly/lambda_datasets}.
 The repository for the code generated during the current study is available at \url{https://github.com/jmflach/SymbolicLambda}

\bibliographystyle{plain}
\bibliography{france.bib}
\end{document}